\documentclass[letterpaper]{article} 
\usepackage{aaai23}  
\usepackage{times}  
\usepackage{helvet}  
\usepackage{courier}  
\usepackage[hyphens]{url}  
\usepackage{graphicx} 
\usepackage{natbib}
\usepackage{caption}
\usepackage{algorithm}
\usepackage{algorithmic}

\usepackage{amssymb}
\usepackage{multirow}
\usepackage{pifont}
\usepackage{makecell}
\usepackage{amsmath}
\usepackage{color}
\usepackage{arydshln}
\usepackage{booktabs}
\usepackage{diagbox}
\usepackage{subcaption}
\definecolor{citecolor}{RGB}{0,0,160}
\definecolor{linkcolor}{RGB}{255,0,0}
\usepackage{hyperref}
\hypersetup{colorlinks=true,citecolor=citecolor,linkcolor=linkcolor}

\usepackage{newfloat}
\usepackage{listings}
\DeclareCaptionStyle{ruled}{labelfont=normalfont,labelsep=colon,strut=off} 
\floatstyle{ruled}
\newfloat{listing}{tb}{lst}{}
\floatname{listing}{Listing}

\title{
DPText-DETR: Towards Better Scene Text Detection with Dynamic Points in Transformer
}
\author {
    Maoyuan Ye\textsuperscript{\rm 1},
    Jing Zhang\textsuperscript{\rm 2},
    Shanshan Zhao\textsuperscript{\rm 3},
    Juhua Liu\textsuperscript{\rm 1}\footnote{Corresponding author. This work was done during Maoyuan Ye’s internship at JD Explore Academy.},
    Bo Du\textsuperscript{\rm 4}\footnotemark[1],
    Dacheng Tao\textsuperscript{\rm 3,2}
}
\affiliations {
    \textsuperscript{\rm 1} Research Center for Graphic Communication, Printing and Packaging, and Institute of Artificial Intelligence, Wuhan University, China \ \ \ \ \ \ \textsuperscript{\rm 2} The University of Sydney, Australia \ \ \ \ \ \
    \textsuperscript{\rm 3} JD Explore Academy, China\\
    \textsuperscript{\rm 4} National Engineering Research Center for Multimedia Software, Institute of Artificial Intelligence, School of Computer Science and Hubei Key Laboratory of Multimedia and Network Communication Engineering, Wuhan University, China\\
    \{yemaoyuan, liujuhua, dubo\}@whu.edu.cn, jing.zhang1@sydney.edu.au, \{sshan.zhao00, dacheng.tao\}@gmail.com
}

\usepackage{bibentry}

\begin{document}

\maketitle

\begin{abstract}
Recently, Transformer-based methods, which predict polygon points or Bezier curve control points for localizing texts, are popular in scene text detection. 
However, these methods built upon detection transformer framework might achieve sub-optimal training efficiency and performance due to coarse positional query modeling. 
In addition, the point label form exploited in previous works implies the reading order of humans, which impedes the detection robustness from our observation.
To address these challenges, this paper proposes a concise Dynamic Point Text DEtection TRansformer network, termed DPText-DETR. In detail, DPText-DETR directly leverages explicit point coordinates to generate position queries and dynamically updates them in a progressive way.
Moreover, to improve the spatial inductive bias of non-local self-attention in Transformer, we present an Enhanced Factorized Self-Attention module which provides point queries within each instance with circular shape guidance. Furthermore, we design a simple yet effective positional label form to tackle the side effect of the previous form. To further evaluate the impact of different label forms on the detection robustness in real-world scenario, we establish an Inverse-Text test set containing 500 manually labeled images. 
Extensive experiments prove the high training efficiency, robustness, and state-of-the-art performance of our method on popular benchmarks. 
The code and the Inverse-Text test set are available at \url{https://github.com/ymy-k/DPText-DETR}.
\end{abstract}

\section{Introduction}
\label{sec:intro}
Text reading and understanding have aroused increasing research interest in the computer vision community~\cite{liao2019mask,liao2020mask,liu2020abcnet,9525302,zhang2020trie,singh2019towards,he2022visual,du2022i3cl,liu2020asts,qiao2021mango,zhou2021tdi}, due to the wide range of practical applications, such as autonomous driving \cite{zhang2020empowering}. To achieve it, as a prerequisite, scene text detection has been studied extensively. However, the distinction of scene text, \emph{e.g.}, complex styles and arbitrary shapes make detection remain challenging.

\begin{figure}[t!]
    \centering
    \subcaptionbox{}{\includegraphics[width=\linewidth]{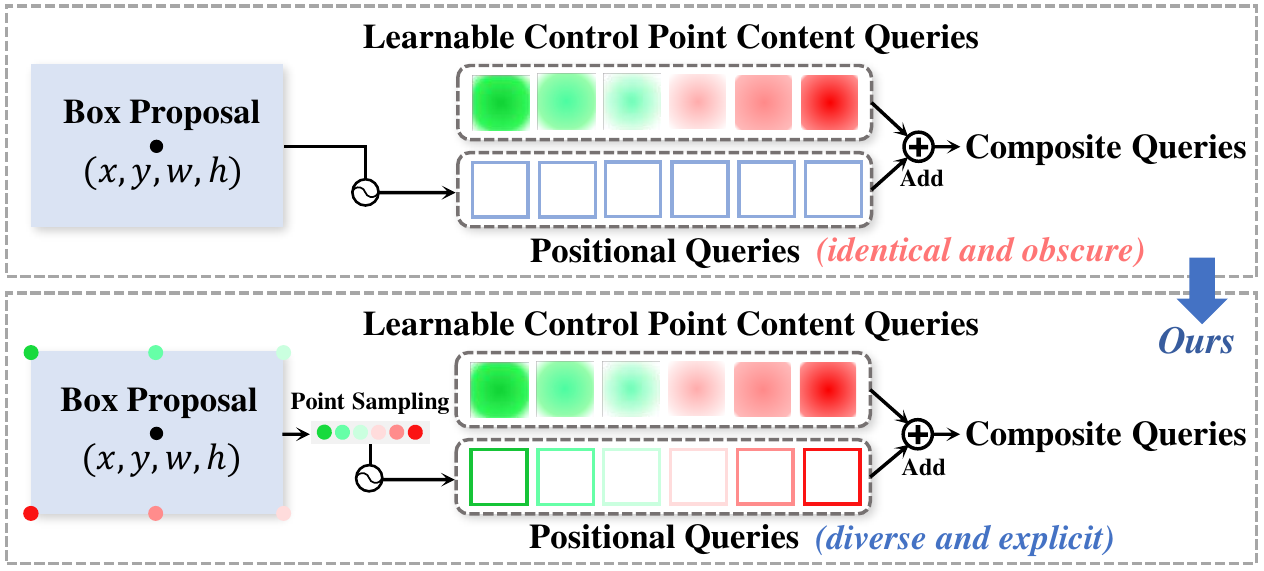}}
    \subcaptionbox{}{\includegraphics[width=2.8cm]{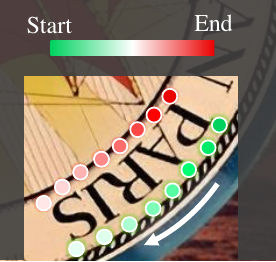}}
    \subcaptionbox{}{\includegraphics[width=2.6cm]{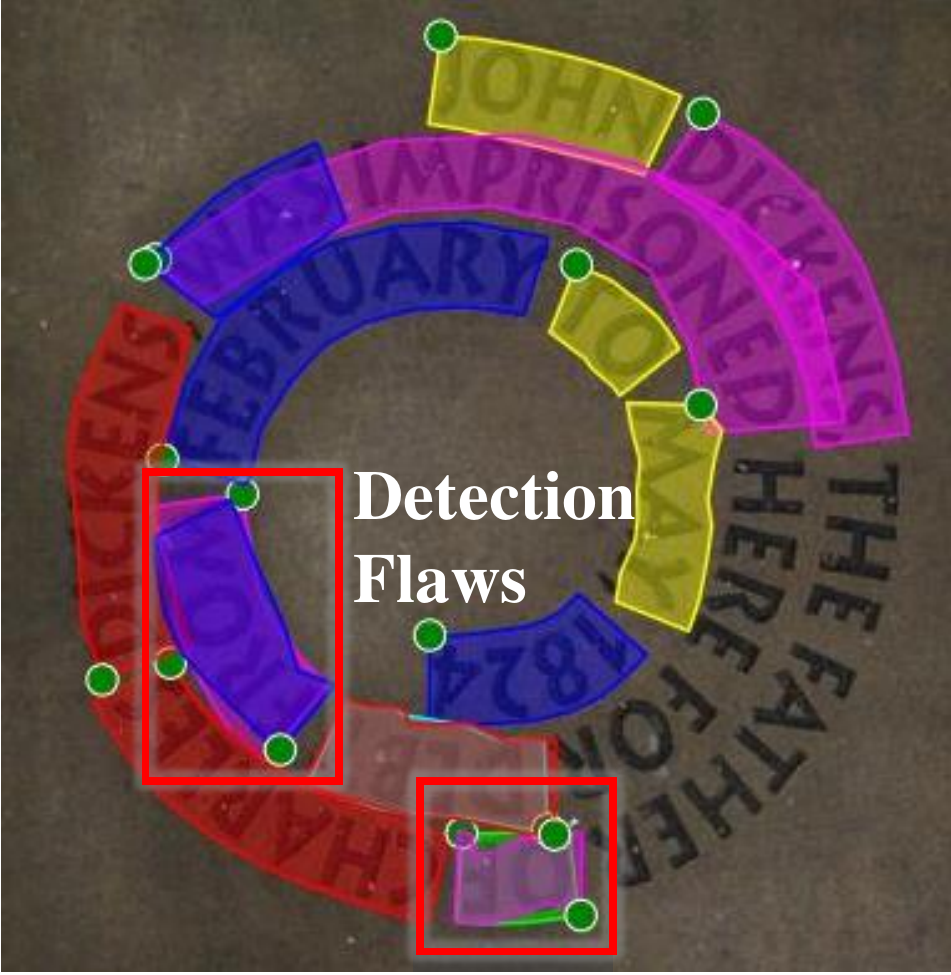}}
    \subcaptionbox{}{\includegraphics[width=2.6cm]{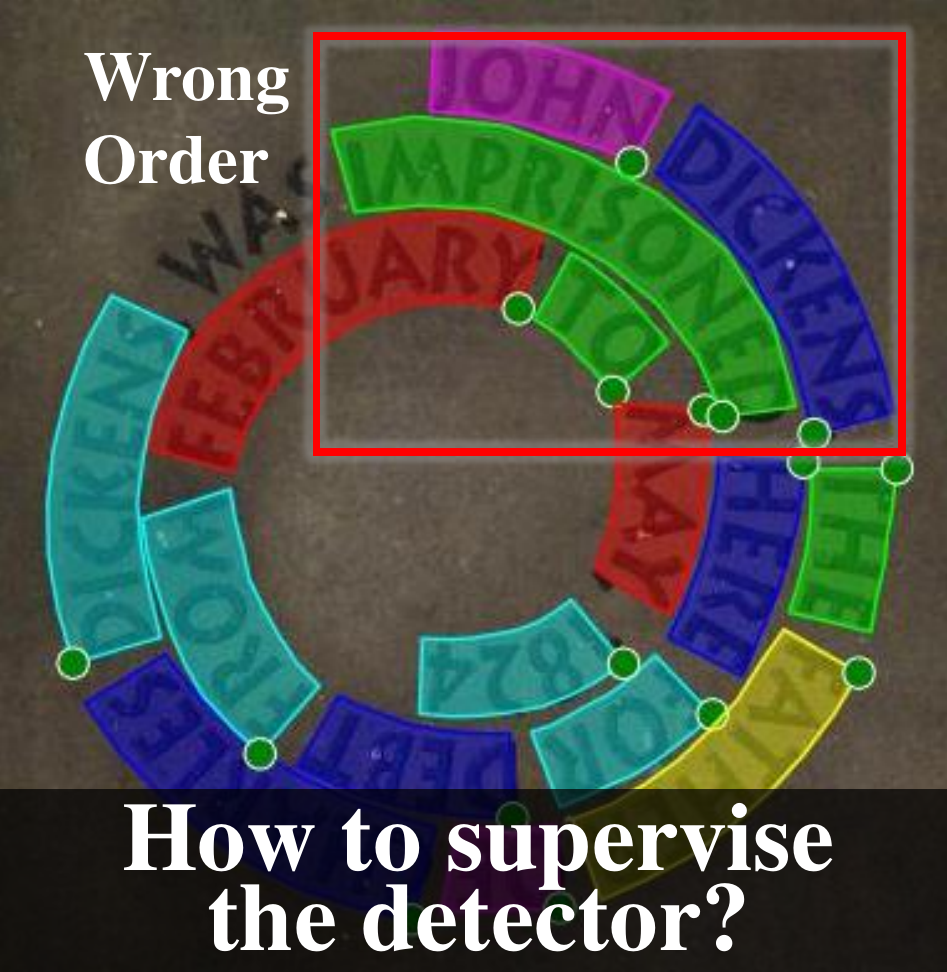}}
    \caption{(a) Comparison of coarse (top) and our explicit (bottom) positional query modeling. (b) The original label implies the reading order of humans. (c) The original label induces the detector to implicitly learn the reading order, resulting in some flaws, such as false positives. Green points are the predicted start points of clockwise reading order. (d) The detector cannot learn the reading order well even with extensive rotation augmentation.}
    \label{fig:Introduction}
\end{figure}

Recently, DETR \cite{carion2020end} introduces Transformer \cite{vaswani2017attention} to object detection, forming a concise and seminal end-to-end framework. Following DETR, lots of works \cite{zhu2020deformable,dai2021dynamic,wang2021fp,meng2021conditional,liu2022dabdetr,wang2022towards} further improve the training convergence and performance. For example, DAB-DETR \cite{liu2022dabdetr} offers insights on the query which can be formed by a content part and a positional part, and proves that the positional part is essential for the training convergence. However, the above detection transformers predicting axis-aligned boxes fall short in handling arbitrary-shape scene texts. In response, recent DETR-like methods \cite{zhang2022text,tang2022few} predict polygon control points or Bezier curve control points following ABCNet \cite{liu2020abcnet}. Specifically, TESTR \cite{zhang2022text} enables Deformable DETR \cite{zhu2020deformable} to predict polygon results in a subtle manner. TESTR uses anchor box proposals from the Transformer encoder to generate positional queries and provide position prior for control point content queries, as shown in the top part of Fig.~\ref{fig:Introduction}(a). However, the position prior from box information is coarse and mismatches the target of predicting points for detection to some extent, which impacts the training efficiency.
We abbreviate it as the \textbf{query formulation issue}. 

In addition, although the scheme of predicting control points enables novel solutions to scene text detection, it also introduces an issue \emph{w.r.t} the order of the points. 
In detail, previous related works adopt the label form of control points according to the reading order of human, as shown in Fig.~\ref{fig:Introduction}(b). This scheme is straightforward, while we are curious about whether it is necessary to enable the detector to localize the text as the human does for understanding the text. Previous works do not 
investigate the impact of such control point label form. However, interestingly, we find this form harms the detection robustness when there are inverse-like texts in the training dataset, even though the ratios of such texts are quite low, \emph{e.g.}, about 2.8$\%$ in Total-Text \cite{ch2020total}, 5.2$\%$ in CTW1500 \cite{liu2019curved}, and 5.3$\%$ in ICDAR2019 ArT \cite{chng2019icdar2019}.  We denote it as the \textbf{label form issue}.
Some detection flaws caused by this issue are shown in Fig.~\ref{fig:Introduction}(c). 
Since there are few inverse-like texts in existing benchmarks, we collect an Inverse-Text test set to further investigate the impact of this label form on detection robustness in real-world scenario. The collected dataset contains 500 scene images with about 40$\%$ inverse-like texts. We hope Inverse-Text can inspire and facilitate future researches through the preliminary attempt in data by filling the gap of lacking inverse-like texts in existing test sets.

To address the query formulation issue and the label form issue, we propose a novel Dynamic Point Text DEtection TRansformer network termed DPText-DETR. In terms of the query formulation issue, we propose an Explicit Point Query Modeling (EPQM) method. Specifically, instead of using boxes, we directly utilize point coordinates to get positional queries, as illustrated in the bottom part of Fig.~\ref{fig:Introduction}(a). With the explicit and complete point formulation, the model is able to dynamically update points in decoder layers. 
Moreover, non-local self-attention lags behind convolution in capturing spatial inductive bias. Hence, we propose an Enhanced Factorized Self-Attention (EFSA) module leveraging circular convolution \cite{peng2020deep} to explicitly model the circular form of polygon points and complement the pure self-attention.
In terms of the label form issue, we design a practical positional label form, which makes the start points independent of the semantic content of texts. With this simple operation, it can noticeably improve the detection robustness.

Overall, our main contributions are three-fold:
\begin{itemize}
\item We propose DPText-DETR, which improves the training convergence and the spatial inductive bias of self-attention by exploiting EPQM and EFSA modules.
\item We investigate the impact of control point label form and design a practical positional label form to improve the detection robustness. We also establish a novel Inverse-Text test set to fill the gap of lacking inverse-like texts in existing datasets.
\item DPText-DETR sets new state-of-the-art on representative arbitrarily-shaped scene text detection benchmarks. It also has fast convergence and promising data efficiency. 
\end{itemize}

\section{Related Work}
\subsection{Detection Transformers}
Transformer \cite{vaswani2017attention} originates from machine translation and soon becomes popular in computer vision community~\cite{dosovitskiy2020image,liu2021swin,xu2021vitae,zhang2022vitaev2,zhang2022vsa}. Recently, the seminal DETR \cite{carion2020end} treats object detection as a set prediction problem and proposes a concise end-to-end framework without complex hand-crafted anchor generation and post-processing. However, DETR suffers from significant slow training convergence and inefficient usage of high resolution features, which have sparked the following researches in detection transformers. For example, Deformable-DETR \cite{zhu2020deformable} attends to sparse features to address above issues. DE-DETR \cite{wang2022towards} identifies the key factor that affects data efficiency is sparse feature sampling. 
DAB-DETR \cite{liu2022dabdetr} uses dynamic anchor boxes as position queries in Transformer decoder to facilitate training. In comparison, in our study, we recast the query in point formulation to handle arbitrary-shape scene texts and speed up training.

\begin{figure*}[!ht]
    \centering
    \includegraphics[width=\linewidth]{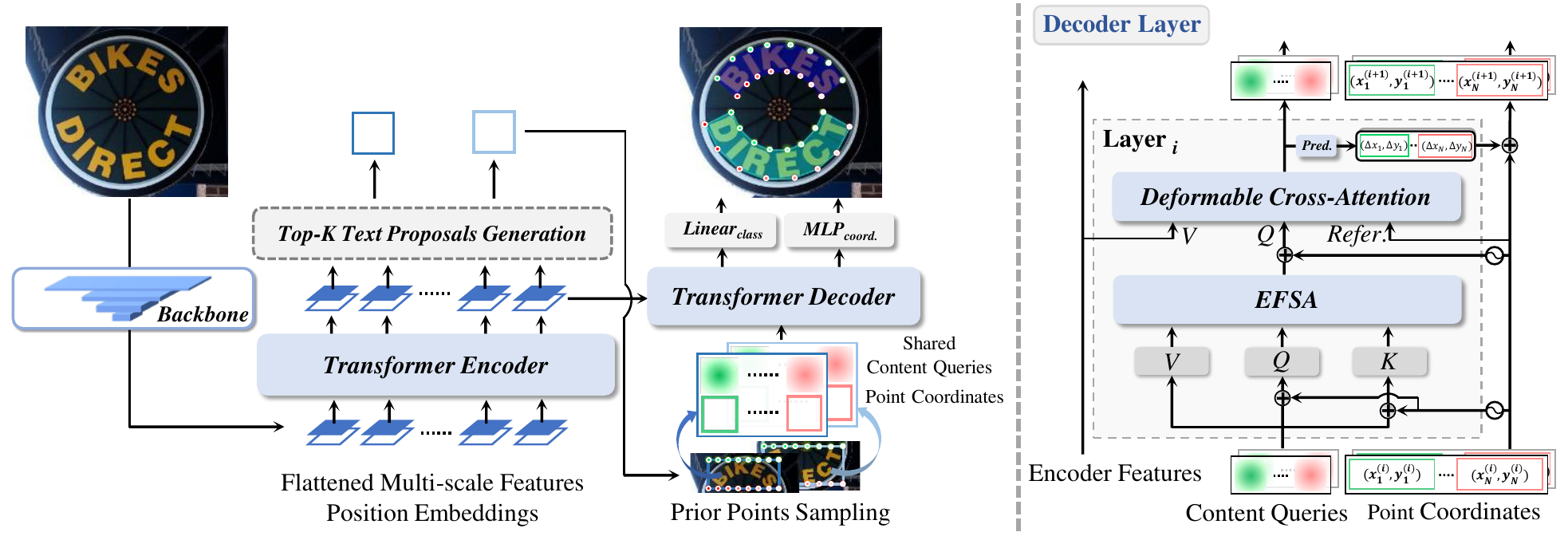}
    \caption{The architecture of DPText-DETR, which mainly consists of a CNN backbone, a Transformer encoder and decoder. Explicit points are calculated by Prior Points Sampling and encoded into positional queries. The point coordinates are progressively refined to form the final polygon predictions.}
    \label{fig:model}
\end{figure*}

\subsection{Contour-based Text Detection and Spotting}
From the perspective of modeling text contour, ABCNet \cite{liu2020abcnet} predicts Bezier curve control points to adaptively fit arbitrarily-shaped texts for the first time. To enhance the capability to localize highly-curved texts, FCENet \cite{zhu2021fourier} models text instances with  Fourier contour fitting. In contrast, TextBPN and its extension \cite{zhang2021adaptive,zhang2022arbitrary} segment various probability maps and use them as priors to generate coarse boundary proposals, then iteratively refine boundary points with graph convolution or Transformer encoder. Considering that segmentation might be sensitive to noise, PCR \cite{dai2021progressive} proposes to progressively evolve the initial text proposal to arbitrarily shaped contours in a top-down manner in the convolution framework. More recently, inspired by detection transformers, FSG\cite{tang2022few} samples a few representative features and uses Transformer encoder layers to implicitly group them, then predicts Bezier curve control points for localization. In comparison, TESTR \cite{zhang2022text} proposes a box-to-polygon scheme for text contour modeling, which utilizes box proposals from Transformer encoder as position queries to guide learnable control points content queries. We conjecture that the box information is coarse for point target in detection, which hinders efficient training. Hence, our work investigates the explicit and complete point query formulation in detection transformer framework.

\section{Methodology}
\label{sec:method}
This paper studies the scene text detection problem by developing an efficient Transformer-based decoder and investigating the influence of control point labels on the detection robustness.
In more details, we propose an Explicit Point Query Modeling (\emph{i.e.}, EPQM, including Prior Points Sampling and Point Update) method and an Enhanced Factorized Self-Attention (EFSA) module in our scene text detector. In this section, we first briefly describe the overall pipeline and then detail the implementation.

\subsection{Overview}
The overall model architecture, which is built upon Deformable-DETR \cite{zhu2020deformable}, is illustrated in Fig.~\ref{fig:model}. In general, given a scene text image, we use a CNN backbone followed by a Transformer encoder to extract features.
After the final encoder layer, multiple axis-aligned boxes are generated according to the two-stage scheme described in Deformable-DETR. With the center point and scale information of each anchor box, we can directly sample a certain number of initial control point coordinates uniformly on the top and bottom sides. In this way, these point coordinates can be used as suitable reference points for the deformable cross-attention module.
In the decoder, point coordinates are encoded and added to corresponding control point content queries to form composite queries. The composite queries are firstly sent to EFSA to further mine their relative relationships and then fed into the deformable cross-attention module. Then control point coordinate prediction head is adopted to dynamically update the reference points layer-by-layer to better fit arbitrary-shape scene text. Finally, prediction heads are used to generate class confidence scores and $N$ control point coordinates for each text instance. During training, we follow \cite{zhang2022text} to calculate losses for classification and control points. More details are described as follows.

\subsection{Positional Label Form}
\label{pos.label}
The original label form shown in Fig.~\ref{fig:Introduction}(b) is in line with human reading order. However, this form induces detector to implicitly learn the order, which increases the learning burden and confuses the model when the texts are in different orders during training. Moreover, even with sufficient rotation augmentations during training, it is difficult for the detector to correctly predict the reading order from visual features alone, as shown in Fig.~\ref{fig:Introduction}(d).

\begin{figure}[t!]
    \centering
    \includegraphics[width=\linewidth]{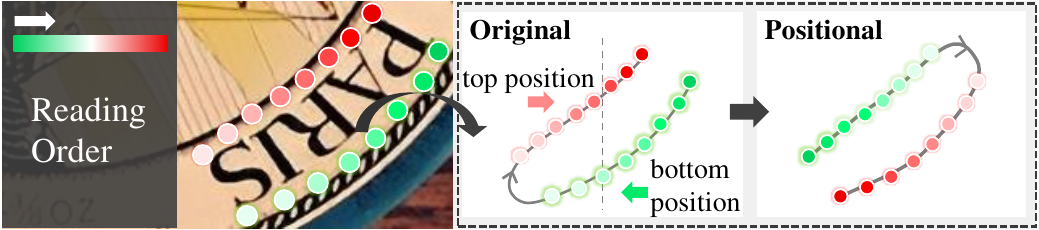}
    \caption{The original and positional label form. The points of each original label are sampled from the Bezier curves \cite{liu2020abcnet} which fit the annotated polygon.}
    \label{fig:ori_pos_label}
\end{figure}

To ease the difficulty, we present a positional label form to guide the detector to distinguish the top and bottom sides of scene text in a pure spatial sense without considering the concrete content of texts. As illustrated in Fig.~\ref{fig:ori_pos_label}, the positional label form mainly follows two simple rules: clockwise order and independent of text content. Specifically, we make the order of all original point labels in clockwise. If the original top side of text instance lies in the bottom position, the starting point is adjusted to the other side. When two sides are arranged left and right, if there is one side with a smaller minimum $y$ value (origin in left-top), the starting point is adjusted to this side, otherwise, it is on the fixed default side.

\subsection{Explicit Point Query Modeling}
\subsubsection{Prior Points Sampling.} It is remarkable to transform axis-aligned box predictions into polygons that fit scene text with a concise yet effective operation, \emph{i.e.}, the box-to-polygon scheme proposed by TESTR \cite{zhang2022text}. Here, we briefly review this scheme. Concretely, after the final encoder layer, each anchor box provided by a top-$K$ proposals generator is encoded, and then shared by $N$ control point content queries. The resulting composite queries $Q^{(i)} (i = 1, \ldots , K)$ can be formulated as follows:
\begin{equation}
Q^{(i)} = P^{(i)} + C = \varphi ((x, y, w, h)^{(i)}) + (p_1, \ldots , p_N),\label{eq_1}
\end{equation}
where $P$ and $C$ represent the positional and the content part of each composite query, respectively. $\varphi$ is the \textit{sine} positional encoding function followed with a linear and normalization layer. $(x, y, w, h)$ represents the center coordinate and scale information of each anchor box. $(p_1, \ldots , p_N)$ is the $N$ learnable control point content queries shared across $K$ composite queries. Note that we set the detector with the query formulation in Eq. (\ref{eq_1}) as our baseline. From Eq. (\ref{eq_1}), we can find that different control point content queries share the same anchor box prior information in each instance. Although the prior facilitates the prediction of control point positions, it mismatches the point targets to some extent. Content queries lack respective explicit position priors to exploit in the box sub-region. 

Motivated by the positional label form and the shape prior that the top and bottom side of a scene text are usually close to the corresponding side on bounding box, we sample $\frac{N}{2}$ point coordinates $point_n(n = 1, \ldots , N)$ uniformly on the top and bottom side of each anchor box, respectively:
\begin{equation}
point_n = \left\{
\begin{array}{cc}
(x - \frac{w}{2} + \frac{(n - 1) \times w}{\frac{N}{2} - 1}, y - \frac{h}{2}), & {n \leq \frac{N}{2}} \\
(x - \frac{w}{2} + \frac{(N - n) \times w}{\frac{N}{2} - 1}, y + \frac{h}{2}), & {n \textgreater \frac{N}{2}}
\end{array} \right. .
\end{equation}
\label{eq_2}
With $(point_1, \ldots , point_N)$, we can generate composite queries using the following complete point formulation: 
\begin{equation}
Q^{(i)} = \varphi' ((point_1, \ldots , point_N)^{(i)}) + (p_1, \ldots , p_N).\label{eq_3}
\end{equation}

In this way, $N$ control point content queries enjoy their respective explicit position prior, resulting in the superior training convergence.

\subsubsection{Point Update.} With the point coordinates, we can refine point positions layer-by-layer and use the updated positions as new reference points for deformable cross-attention. In comparison, TESTR directly adopts the anchor boxes information to generate position queries. Therefore, it is hard to perform refinement between decoder layers. Specifically, in our model, we update control points in each decoder layer after getting respective offsets $(\Delta x, \Delta y)$ by a prediction head, as illustrated in the decoder layer part of Fig.~\ref{fig:model}.

 \subsubsection{Discussion.} We also notice that the very recent work BoundaryFormer \cite{lazarow2022instance} adopts a similar point query formulation in the instance segmentation task. BoundaryFormer aims at predicting polygons yet uses instance mask supervision. In BoundaryFormer, a fully convolutional detector \cite{ren2015faster,tian2019fcos} is exploited to predict object boxes. Next, a diamond is initialized for each box. Then, a Transformer decoder is used to refine the position of vertexes. Between decoder layers, new points are inserted between existing ones to produce fine polygons. In comparison, we aim to address the training concerned issue by modeling explicit and complete point queries. In our model, a fixed number of points are sampled on the top and bottom sides of each proposal box before the Transformer decoder, according to the shape prior that texts can be located with only two sides. The explicit point formulation enables the decoder to iteratively refine points for more precise final predictions in both BoundaryFormer and our model. However, in our DETR-based model, the explicit point query formulation is further explored to address the relatively slow convergence issue.

\subsection{Enhanced Factorized Self-Attention}
Following \cite{zhang2022text}, we exploit Factorized Self-Attention (FSA) \cite{dong2021visual} in our baseline. In FSA, an intra-group self-attention ($SA_{intra}$) across $N$ subqueries belonging to each of the $Q^{(i)}$ is firstly exploited to capture the relationship between different points within each text instance. After $SA_{intra}$, an inter-group self-attention ($SA_{inter}$) across $K$ composite queries is adopted to capture the relationship between different instances. We conjecture that the non-local $SA_{intra}$ falls short in capturing the circular shape prior of polygon control points. Hence, we leverage the local circular convolution \cite{peng2020deep} to complement FSA, forming the Enhanced Factorized Self-Attention. Concretely, $SA_{intra}$ is firstly performed to get queries $Q_{intra} = SA_{intra}(Q)$, where keys are the same as $Q$ while values exclude the positional part. Meanwhile, locally enhanced queries are generated: $Q_{local} = ReLU(BN(CirConv(Q)))$. Then, fused queries can be obtained: $Q_{fuse} = LN(FC(C + LN(Q_{intra} + Q_{local})))$, where $C$ represents the content queries used as a shortcut, $FC$ is a fully connected layer, $BN$ is BatchNorm, and $LN$ is LayerNorm. Next, the relationships between different instances are mined: $Q_{inter} = SA_{inter}(Q_{fuse})$. After that, $Q_{inter}$ is sent to the deformable cross-attention module. Using one circular convolution layer with four-neighborhood achieves the best trade-off between performance and inference speed. We adopt this setting for experiments.

\section{Experiments}
\label{sec:exp}
We conduct experiments on three arbitrary-shape scene text benchmarks: Total-Text \cite{ch2020total}, CTW1500 \cite{liu2019curved} and ICDAR19 ArT \cite{chng2019icdar2019}. Ablation studies are conducted on Total-Text to verify the effectiveness of each component of our methods.

\begin{table*}\normalsize
\centering
\setlength{\tabcolsep}{10pt}
\resizebox{0.9\hsize}{!}{
\begin{tabular}{l l ccc ccc ccc}
\toprule[1.5pt]
\multirow{2}{*}{Method} & \multirow{2}{*}{Backbone} & \multicolumn{3}{c}{\textbf{Total-Text}} & \multicolumn{3}{c}{\textbf{CTW1500}} & \multicolumn{3}{c}{\textbf{ICDAR19 ArT}} \\
\cmidrule(lr){3-5} \cmidrule(lr){6-8} \cmidrule(lr){9-11}
& & P & R & F & P & R & F & P & R & F \\
\midrule[1.1pt]
TextSnake~\cite{long2018textsnake} & VGG16  & 82.7 & 74.5 & 78.4     & 67.9 & 85.3 & 75.6    & $-$ & $-$ & $-$ \\
PAN~\cite{wang2019efficient} & Res-18  & 89.3 & 81.0 & 85.0     & 86.4 & 81.2 & 83.7    & $-$ & $-$ & $-$ \\
CRAFT~\cite{baek2019character} \dag & VGG16  & 87.6 & 79.9 & 83.6     & 86.0 & 81.1 & 83.5    & 77.2 & 68.9 & 72.9 \\
TextFuseNet~\cite{ye2020textfusenet} \dag & Res50  & 87.5 & 83.2 & 85.3     & 85.8 & 85.0 & 85.4    & 82.6 & 69.4 & 75.4 \\
DB~\cite{liao2020real} & Res50-DCN  & 87.1 & 82.5 & 84.7     & 86.9 & 80.2 & 83.4    & $-$ & $-$ & $-$ \\
PCR~\cite{dai2021progressive} & DLA34  & 88.5 & 82.0 & 85.2     & 87.2 & 82.3 & 84.7    & 84.0 & 66.1 & 74.0 \\
ABCNet-v2~\cite{9525302} & Res50  & 90.2 & 84.1 & 87.0     & 85.6 & 83.8 & 84.7    & $-$ & $-$ & $-$ \\
I3CL~\cite{du2022i3cl} & Res50  & 89.2 & 83.7 & 86.3     & 87.4 & 84.5 & 85.9    & 82.7 & 71.3 & \underline{76.6} \\
TextBPN++~\cite{zhang2022arbitrary} & Res50  & 91.8 & 85.3 & \underline{88.5}     & 87.3 & 83.8 & 85.5    & 81.1 & 71.1 & 75.8 \\
FSG~\cite{tang2022few} & Res50  & 90.7 & 85.7 & 88.1     & 88.1 & 82.4 & 85.2    & $-$ & $-$ & $-$ \\
TESTR-polygon~\cite{zhang2022text} & Res50  & 93.4 & 81.4 & 86.9     & 92.0 & 82.6 & 87.1    & $-$ & $-$ & $-$ \\
SwinTextSpotter~\cite{huang2022swintextspotter} & Swin  & $-$ & $-$ & 88.0     & $-$ & $-$ & \underline{88.0}    & $-$ & $-$ & $-$ \\
\midrule
DPText-DETR (ours) & Res50   & 91.8 & 86.4 & \textbf{89.0}   & 91.7 & 86.2 & \textbf{88.8}   & 83.0 & 73.7 & \textbf{78.1} \\
\bottomrule[1.5pt]
\end{tabular}
}
\caption{Quantitative detection results on benchmarks. ``P'', ``R'' and ``F'' denote Precision, Recall and F-measure, respectively. F-measure is the major evaluation metric. ``\dag'' means that the results on ICDAR19 ArT are collected from the official website \cite{chng2019icdar2019}.}
\label{tab:main_results}
\end{table*}

\subsection{Datasets}
First, we briefly introduce the exploited datasets. \textbf{SynthText 150K} \cite{liu2020abcnet} is a synthesized dataset for arbitrary-shape scene text, containing 94,723 images with multi-oriented text and 54,327 images with curved text. \textbf{Total-Text} \cite{ch2020total} consists of 1,255 training images and 300 test images. Word-level polygon annotations are provided. \textbf{Rot.Total-Text} is a test set derived from the Total-Text test set. Since the original label form induces model to generate unstable prediction as shown in Fig.~\ref{fig:Introduction}(c), we apply large rotation angles ($45^{\circ}$, $135^{\circ}$, $180^{\circ}$, $225^{\circ}$, $315^{\circ}$) on images of the Total-Text test set to examine the model robustness, resulting in 1,800 test images including the original test set. \textbf{CTW1500} \cite{liu2019curved} contains 1,000 training images and 500 test images. Text-line level annotations are presented. \textbf{ICDAR19 ArT} \cite{chng2019icdar2019} is a large arbitrary-shape scene text benchmark. It contains 5,603 training images and 4,563 test images. 

\textbf{Inverse-Text} established in our work, consists of 500 test images. It is a arbitrary-shape scene text test set with about 40$\%$ inverse-like instances. A few instances are mirrored due to photographing. Some images are selected from existing benchmark test sets, \emph{i.e.}, 121 images from ICDAR19 ArT, 7 images from Total-Text, and 3 images from CTW1500. Other images are collected from the Internet. Word-level polygon annotations are provided. Some samples are shown in Fig.~\ref{fig:vis_inversetext}. 

\begin{figure}[!t]
    \centering
    \includegraphics[width=\linewidth]{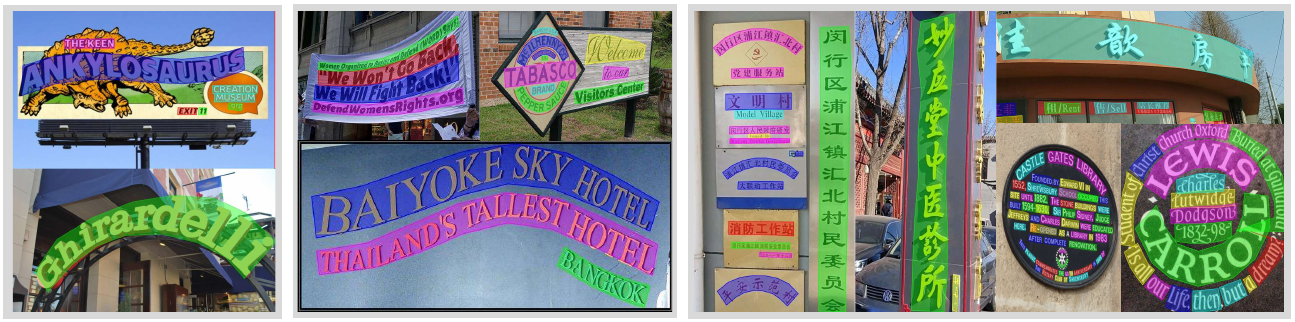}
    \caption{Qualitative results on Total-Text, CTW1500, and ICDAR19 ArT, from left to right.}
    \label{fig:qualitative_tt_ctw_art}
\end{figure}

\subsection{Implementation Details}
\label{subsec:impl details}
We adopt ResNet-50 \cite{he2016deep} as the backbone. We use 8 heads for multi-head attention and 4 sampling points for deformable attention. The number of both encoder and decoder layers is set to 6. The composite queries number $K$ is 100 and default control points number $N$ is 16. We follow the hyper-parameter setting of loss used in the detection part of \cite{zhang2022text}. Models are trained with 4 NVIDIA A100 (40GB) GPUs and tested with 1 GPU.

In ablation studies, we do not pre-train models to intuitively reveal the training convergence on Total-Text. We train models on Total-Text for 120k without rotation data augmentation and directly test them on Rot.Total-Text and Inverse-Text to verify the robustness. To help the model adapt to different text orders, we additionally rotate Total-Text training images with six angles ($-45^{\circ}$, $-30^{\circ}$, $-15^{\circ}$, $15^{\circ}$, $30^{\circ}$, $45^{\circ}$) representing normal cases, and rotate all normal cases for $180^{\circ}$ representing inverse cases. When using rotation data, we train models for 200k iterations.

The complete training process is divided into two stages: pre-training stage and finetuning stage. The batch size is set to 8. For Total-Text and CTW1500, following \cite{zhang2022text,9525302}, the detector is pre-trained on a mixture of SynthText 150K, MLT \cite{nayef2019icdar2019} and Total-Text for 350k iterations. The initial learning rate ($lr$) is $1 \times 10^{-4}$ and is decayed to $1 \times 10^{-5}$ at 280k. We finetune it on Total-Text for 20k iterations, with $5 \times 10^{-5}$ $lr$ which is divided by 10 at 16k. We adopt 13k finetuning iterations for CTW1500, with $2 \times 10^{-5}$ $lr$. For ICDAR19 ArT, following \cite{du2022i3cl,baek2020character}, we adopt LSVT \cite{sun2019icdar} during pre-training. We use a mixture of SynthText 150K, MLT, ArT and LSVT to pre-train the model for 400k iterations. $lr$ is $1 \times 10^{-4}$ and is decayed to $1 \times 10^{-5}$ at 320k. Then, we finetune it on ArT for 50k iterations, with $5 \times 10^{-5}$ $lr$ which is divided by 10 at 40k. We use the AdamW optimizer \cite{Loshchilov2019DecoupledWD} with $\beta_1=0.9$, $\beta_2=0.999$ and weight decay of $10^{-4}$. Data augmentation strategies such as random crop, random blur, brightness adjusting, and color change are applied. Note that rotation data mentioned above is only used in finetuning stage for each benchmark. We adopt multi-scale training strategy with the shortest edge ranging from 480 to 832, and the longest edge kept within 1600.

\begin{table*}[t!]
    \centering
    \setlength{\tabcolsep}{10pt}
    \resizebox{0.7\linewidth}{!}{
    \begin{tabular}{lccccccccccc}
         \toprule[1.5pt]
         \multirow{2}{*}{\emph{ID}} &
         \multirow{2}{*}{Method} &
         \multirow{2}{*}{Pos.Label} &
         \multirow{2}{*}{EPQM} &
         \multirow{2}{*}{EFSA} &
         \multirow{2}{*}{Rotation} &
         \multicolumn{2}{c}{\textbf{Total-Text}} &
         \multicolumn{2}{c}{\textbf{Rot.Total-Text}} &
         \multicolumn{2}{c}{\textbf{Inverse-Text}} \\
         \cmidrule(lr){7-8} \cmidrule(lr){9-10} \cmidrule(lr){11-12}
         &&&&&& F & FPS & F & FPS & F & FPS \\
         \midrule[1.1pt]
         1 &\multirow{6}{*}{$Baseline$} & & & & &83.90 &\textbf{18.5} &70.02 &\textbf{20.2} &77.63 &\textbf{18.9} \\
         2 & &\checkmark & & & &84.58 &18.5 &73.92 &20.2 &80.70 &18.9 \\
         3 & & &\checkmark & & &85.15 &17.9 &71.28 &19.7 &79.44 &18.5 \\
         4 & & &\checkmark &\checkmark & &85.86 &17.3 &72.60 &18.9 &80.33 &18.3 \\
         5 & &\checkmark &\checkmark & & &85.28 &17.9 &74.87 &19.7 &81.56 &18.5 \\
         6 & &\checkmark &\checkmark &\checkmark & &\textbf{86.17} &17.3 &\textbf{74.99} &18.9 &\textbf{81.99} &18.3 \\
         \midrule
         7 &\multirow{6}{*}{$Baseline$} & & & &\checkmark &84.98 &\textbf{18.5} &83.99 &\textbf{20.2} &84.28 &\textbf{18.9} \\
         8 & &\checkmark & & &\checkmark &86.07 &18.5 &84.52 &20.2 &86.69 &18.9 \\
         9 & & &\checkmark & &\checkmark &86.16 &17.9 &84.15 &19.7 &83.79 &18.5 \\
         10 & & &\checkmark &\checkmark &\checkmark &86.21 &17.3 &84.53 &18.9 &85.95 &18.3 \\
         11 & &\checkmark &\checkmark & &\checkmark &86.46 &17.9 &84.86 &19.7 &86.57 &18.5 \\
         12 & &\checkmark &\checkmark &\checkmark &\checkmark &\textbf{86.79} &17.3 &\textbf{84.95} &18.9 &\textbf{86.78} &18.3 \\
         \bottomrule[1.5pt]
    \end{tabular}}
    \caption{Ablations on test sets. ``Pos.Label'' denotes the positional label form. Without EFSA means the FSA is used instead.}
    \label{tab:main ablation}
\end{table*}

\subsection{Comparison with State-of-the-art Methods}
We test our method on Total-Text, CTW1500, and ICDAR19 ArT. Quantitative results compared with previous methods are presented in Tab.~\ref{tab:main_results}. Our method achieves consistent state-of-the-art performance. Compared with other detectors, for example, DPText-DETR outperforms TexBPN++ by 0.5\%, 3.3\%, and 2.3\% in terms of F-measure on Total-Text, CTW1500, and ICDAR2019 ArT, respectively. Moreover, DPText-DETR leads I3CL by 2.7\%, 2.9\%, and 1.5\% F-measure on the three benchmarks. Compared with FSG, our method achieves 0.9\% and 3.6\% higher F-measure on Total-Text and CTW1500. DPText-DETR also outperforms the state-of-the-art SwinTextSpotter by 1.0\% and 0.8\% in terms of F-measure on Total-Text and CTW1500. 
Some visual results are provided in Fig.~\ref{fig:qualitative_tt_ctw_art}. It shows that DPText-DETR performs well on straight, curve, and even dense long texts. A failure case is also shown, \emph{i.e.}, the right bottom image in ICDAR19 ArT, where the polygon prediction is affected by extremely compact curved texts. 

\subsection{Ablation Studies}
As mentioned before, pre-training is not used in all experiments of this subsection. Main ablation results are reported in Tab.~\ref{tab:main ablation}. Notably, compared with previous pre-trained models, DPText-DETR without pre-training can still achieve competitive performance (F-measure: 86.79$\%$).

\begin{table}[t!]
    \centering
    \setlength{\tabcolsep}{4pt}
    \resizebox{\hsize}{!}{
    \begin{tabular}{lccccccccc}
         \toprule[1.5pt]
         \multirow{2}{*}{Method} &
         \multirow{2}{*}{Pos.Label} &
         \multirow{2}{*}{EPQM} &
         \multirow{2}{*}{Rotation} &
         \multicolumn{2}{c}{\textbf{Total-Text}} &
         \multicolumn{2}{c}{\textbf{Rot.Total-Text}} &
         \multicolumn{2}{c}{\textbf{Inverse-Text}} \\
         \cmidrule(lr){5-6} \cmidrule(lr){7-8} \cmidrule(lr){9-10}
         &&& &F & FPS & F & FPS & F & FPS \\
         \midrule[1.1pt]
         \multirow{3}{*}{$Baseline$} & & & &84.71 &\textbf{18.0} &71.37 &\textbf{19.8} &78.71 &\textbf{20.0} \\
         &\checkmark & & &84.77 &18.0 &\textbf{74.11} &19.8 &\textbf{81.20} &20.0 \\
         &\checkmark &\checkmark & &\textbf{85.80} &17.0 &73.82 &19.0 &80.86 &18.9 \\
         \midrule
         \multirow{3}{*}{$Baseline$} & & &\checkmark &85.83 &\textbf{18.0} &84.16 &\textbf{19.8} &84.86 &\textbf{20.0} \\
         &\checkmark & &\checkmark &86.30 &18.0 &84.65 &19.8 &85.73 &20.0 \\
         &\checkmark &\checkmark &\checkmark &\textbf{86.64} &17.0 &\textbf{85.03} &19.0 &\textbf{86.20} &18.9 \\
         \bottomrule[1.5pt]
    \end{tabular}}
    \caption{Main ablation results of Bezier control point version detector on Total-Text, Rot.Total-Text, and Inverse-Text.}
    \label{tab:ablation_bezier}
\end{table}

\textbf{Positional Label Form.} As shown in Tab.~\ref{tab:main ablation}, when the positional label form is used, the F-measure scores on all test sets are improved. For example, the comparison between the line 1 and line 2 in the table demonstrates that the F-measure is improved by 0.68$\%$ on Total-Text, 3.90$\%$ on Rot.Total-Text and 3.07$\%$ on Inverse-Text, which validates the effectiveness for model robustness. Moreover, positional label form can synergize better with rotation augmentation than the original form to improve the detection performance and robustness. When using rotation, the positional label form also contributes to faster convergence as shown in Fig.~\ref{fig:curves}(a). The positional label is also valid for the model which predicts Bezier control points, as shown in Tab.~\ref{tab:ablation_bezier}.

\textbf{EPQM.} In Tab.~\ref{tab:main ablation}, we investigate the effectiveness of EPQM. EPQM intuitively boosts the performance and makes the major contribution to the convergence as shown in Fig.~\ref{fig:curves}(a). EPQM also improves the performance of Bezier variant on Total-Text consistently, as shown in Tab.~\ref{tab:ablation_bezier}. Moreover, EPQM significantly enhances the few-shot learning ability. As shown in Tab.~\ref{tab:few-shot}, when the training iterations and data volume are decreased, huge performance degradation of baseline models turns up, while the models with EPQM are far less affected.

\begin{table}[t]
    \centering
    \setlength{\tabcolsep}{4pt}
    \resizebox{\hsize}{!}{
    \begin{tabular}{lccccccc}
         \toprule[1.5pt]
         \multirow{2}{*}{Method} &
         \multirow{2}{*}{EPQM} &
         \multirow{2}{*}{EFSA} &
         \multirow{2}{*}{TD-Ratio} &
         \multicolumn{2}{c}{Total-Text} &
         \multicolumn{2}{c}{Inverse-Text} \\
         \cmidrule(lr){5-6} \cmidrule(lr){7-8}
         & & & &F &Improv. &F &Improv. \\
         \midrule[1.1pt]
         \multirow{3}{*}{$Baseline\ w/\ Pos.Label$} & & &100$\%$ &73.92 &$-$ &70.90 &$-$ \\
         &\checkmark & &100$\%$ &82.99 &9.07 &78.18 &7.28 \\
         &\checkmark &\checkmark &100$\%$ &\textbf{83.66} &\textbf{9.74} &\textbf{79.09} &\textbf{8.19} \\
         \midrule
         \multirow{3}{*}{$Baseline\ w/\ Pos.Label$} & & &50$\%$ &30.45 &$-$ &22.62 &$-$ \\
         &\checkmark & &50$\%$ &78.90 &48.45 &72.78 &50.16 \\
         &\checkmark &\checkmark &50$\%$ &\textbf{80.22} &\textbf{49.77} &\textbf{73.97} &\textbf{51.35} \\
         \midrule
         \multirow{3}{*}{$Baseline\ w/\ Pos.Label$} & & &25$\%$ &14.94 &$-$ &6.98 &$-$ \\
         &\checkmark & &25$\%$ &58.54 &43.6 &52.32 &45.34 \\
         &\checkmark &\checkmark &25$\%$ &\textbf{70.49} &\textbf{55.55} &\textbf{60.15} &\textbf{53.17} \\
         \bottomrule[1.5pt]
    \end{tabular}}
    \caption{Fewer iterations and fewer training data test for EPQM and EFSA. ``TD-Ratio'' represents training data ratio compared with the original one. ``Improv.'' represents the improvement on F-measure. In the first three rows, models are only trained for 12k iterations on Total-Text without rotation augmentation and directly tested on Inverse-Text. In the rest parts, we randomly sample training data according to TD-Ratio while keeping the equivalent training epochs as used in the first three rows. We train 6k iterations for the middle three models and 3k iterations for the last ones.}
    \label{tab:few-shot}
\end{table}

\textbf{EFSA.} In Tab.~\ref{tab:main ablation} and Tab.~\ref{tab:few-shot}, we verify the effectiveness of EFSA. The comparison between line 5 and line 6 in Tab.~\ref{tab:main ablation} shows that EFSA can improve the F-measure by 0.89$\%$.
Tab.~\ref{tab:few-shot} shows that EFSA enables the model to learn better with fewer samples. For example, when the training data volume is 25$\%$, compared with the model only equipped with EPQM, the model with both EPQM and EFSA achieves an extra gain of 11.95$\%$ F-measure on Total-Text and 7.83$\%$ F-measure on Inverse-Text. Moreover, as shown in Fig.~\ref{fig:curves}(a), EFSA can further promote the training convergence and the model with all components achieves about six times faster convergence than the baseline in the initial training stage. We find EFSA is more effective when predicting polygon control points. Since Bezier curve control points do not always form in circular shape and sometimes they are far apart, it is not suitable to combine circular convolution with self-attention for the Bezier variant.

In summary, the positional label form mainly improves the model robustness while EPQM and EFSA boost the overall performance, training convergence, and few-shot learning ability. DPText-DETR achieves ideal trade-off between performance gain and inference speed drop. 

\subsection{What Makes Faster Training Convergence?}
\begin{figure}[t!]
    \centering
    \includegraphics[width=\linewidth]{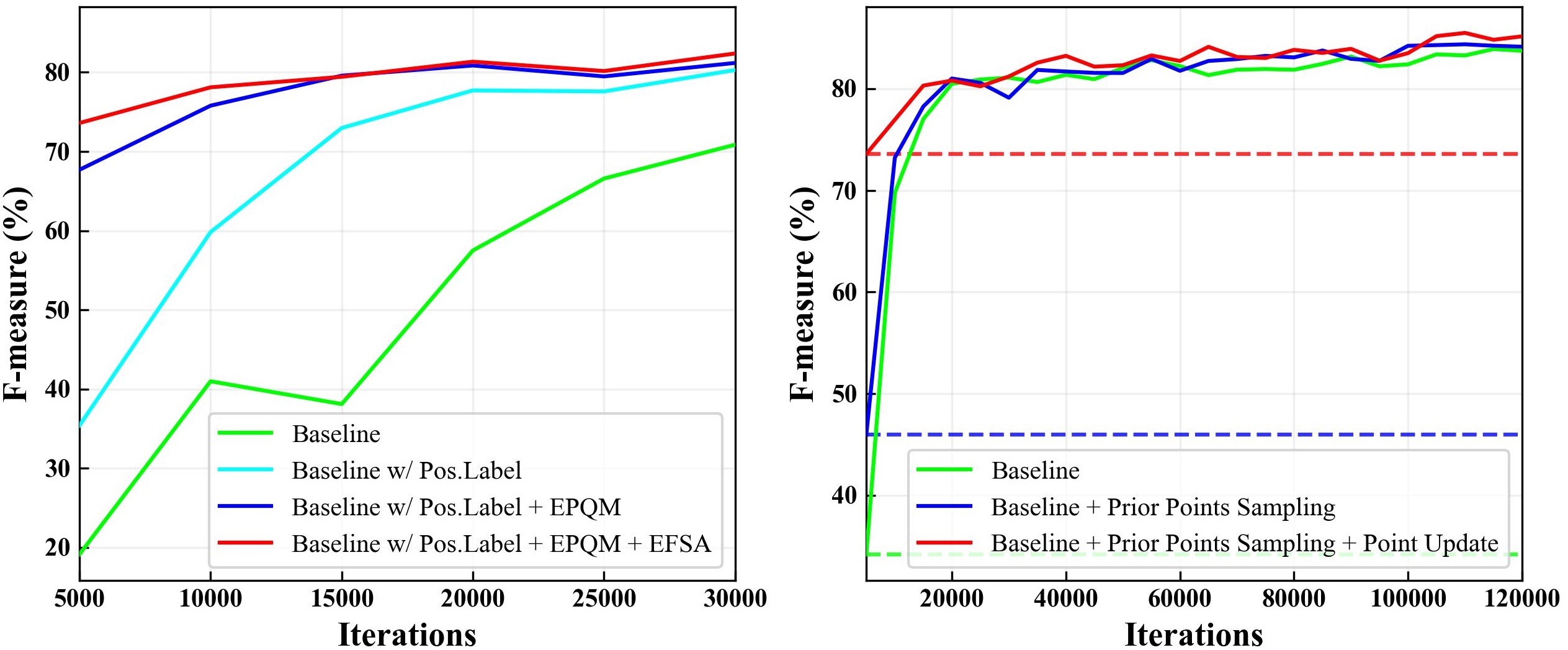}
    \caption{Convergence curves on Rot.Total-Text (left) and Total-Text (right).}
    \label{fig:curves}
\end{figure}

We conduct further ablation studies on EPQM to reveal what makes convergence faster. Quantitative results and convergence curves are shown in Tab.~\ref{tab:analysis on EPQM} and Fig.~\ref{fig:curves}(b). Referring to the blue curve in Fig.~\ref{fig:curves}(b), the convergence at the initial stage is improved when only Prior Points Sampling is used. Referring to the red curve and Tab.~\ref{tab:analysis on EPQM}, Point Update further boosts the convergence by a large margin and makes the major contribution to the performance. It demonstrates that the explicit position modeling for sparse points is the key to faster convergence. The explicit formulation is the prerequisite for dynamically updating points in decoder layers. Dynamic updating provides more precise reference points for deformable cross-attention, resulting in better performance. Prior works \cite{liu2022dabdetr,wang2022towards} have proved that box query formulation and sparse box area features extracted by ROIAlign can improve the training efficiency of DETR-based models. In our DPText-DETR designed for scene text detection, the Prior Points Sampling scheme can be regarded as a soft grid-sample operation, and it is also proved that the point query formulation, which is more sparse than the box, is more beneficial to training. 

\begin{table}[t]
    \centering
    \resizebox{0.8\linewidth}{!}{
    \begin{tabular}{lccc}
         \toprule[1.5pt]
         Method &Prior Points Sampling &Point Update &F \\
         \midrule[1.1pt]
         \multirow{3}{*}{$Baseline$} & & &83.90 \\
         &\checkmark & &84.13 \\
         &\checkmark &\checkmark &\textbf{85.15} \\
         \bottomrule[1.5pt]
    \end{tabular}}
    \caption{Quantitative analysis on EPQM. The results are test on Total-Text without using positional label and EFSA.}
    \label{tab:analysis on EPQM}
\end{table}
\begin{table}[t]
    \centering
    \resizebox{\hsize}{!}{
    \begin{tabular}{lcccc cc}
         \toprule[1.5pt]
         \multirow{2}{*}{Method} &
         \multirow{2}{*}{Rotation} &
         \multirow{2}{*}{P} &
         \multirow{2}{*}{R} &
         \multirow{2}{*}{F} &
         \multicolumn{2}{c}{End-to-End} \\
         \cmidrule(lr){6-7}
         & & & & &None &Full \\
         \midrule[1.1pt]
         ABCNet-v2 (Our repro.) &\checkmark &83.4 &73.2 &78.0 &57.2 &69.5 \\
         ABCNet-v2 w/ Pos.Label (Our repro.) &\checkmark &90.7 &83.9 &87.2 &62.2 &\textbf{76.7} \\
         TESTR (Our repro.) &\checkmark &89.4 &84.4 &86.8 &62.1 &74.7\\
         TESTR w/ Pos.Label (Our repro.) &\checkmark &88.8 &85.7 &87.2 &61.9 &74.1\\
         TESTR w/ Pos.Label (Our detector) &\checkmark &90.7 &84.2 &\underline{87.3} &\textbf{63.1} &\underline{75.4}\\
         SwinTextSpotter (Our repro.) &\checkmark &94.5 &84.7 &\textbf{89.3} &\underline{62.9} &74.7 \\
         \bottomrule[1.5pt]
    \end{tabular}}
    \caption{Results of spotters on Inverse-Text. ``repro.'' and ``None'' indicates our experiment using official released code and the end-to-end results without using lexicon.}
    \label{tab:end2end on inversetext}
\end{table}

\subsection{Further Discussion}
\label{sec:further discussion}
In addition, we further investigate the performance of some arbitrary-shape scene text spotters on Inverse-Text. Recent methods can be roughly categorized into point-based and segmentation-based methods. For point-based methods, we select ABCNet-v2 \cite{9525302} and TESTR \cite{zhang2022text} that exploits dual Transformer decoders for parallel detection and recognition. For segmentation-based methods, we select SwinTextSpotter \cite{huang2022swintextspotter} as a representative. We finetune the official models trained on Total-Text with rotation augmentation as mentioned in implementation details for better adaptation to inverse-like texts. Results are reported in Tab.~\ref{tab:end2end on inversetext}. For ABCNet-v2 and TESTR, we also test the influence of the positional label form. As shown in Tab.~\ref{tab:end2end on inversetext}, the detection F-measures are improved when the positional label form is used, which validates the positive effect on detection. 

\begin{figure}[!]
    \centering
    \includegraphics[width=\linewidth]{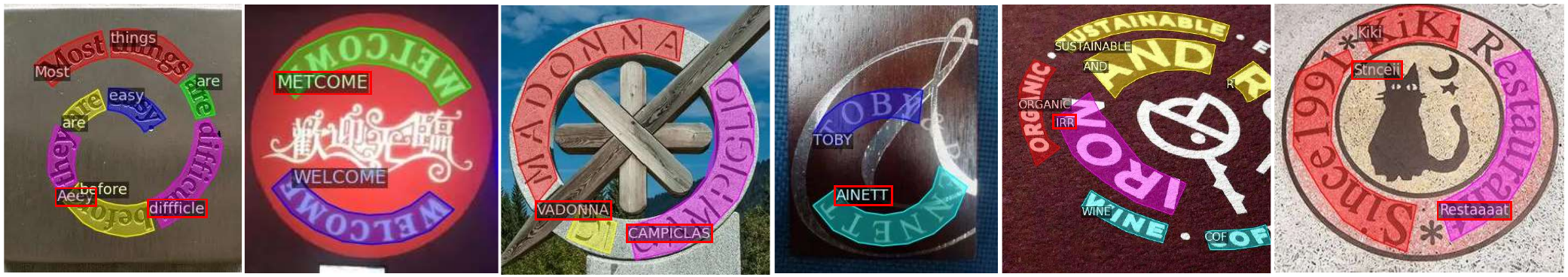}
    \caption{Qualitative results on Inverse-Text. Recognition failures on hard inverse-like texts are marked with red boxes.}
    \label{fig:vis_inversetext}
\end{figure}

We further replace the detection decoder of TESTR with ours, and find that the modified spotter still works well. It indicates that the detection decoder can iteratively refine control points with explicit point information while the recognition decoder remains to learn semantics from a coarse text anchor box sub-region. However, the modified spotter suffers from unsynchronized convergence between detection and recognition. We plan to explore a training efficient Transformer-based spotter in the future. Some visualizations are presented in Fig.~\ref{fig:vis_inversetext}.

\section{Conclusion}
We present a concise yet effective scene text detection transformer network, which transforms composite queries into explicit and complete point formulation. We investigate the effect of control point labels on model robustness and point out a practical positional label form. Extensive experiments demonstrate the state-of-the-art performance, training efficiency, and robustness of our proposed DPText-DETR. We also establish an Inverse-Text test set to facilitate future research in this area.

\section*{Acknowledgements}
This work was supported in part by the National Natural Science Foundation of China under Grants 62076186, 62141112, and 62225113, and in part by the Science and Technology Major Project of Hubei Province (Next-Generation AI Technologies) under Grant 2019AEA170. Dr. Jing Zhang is supported by the ARC project FL-170100117. The numerical calculations in this paper have been done on the supercomputing system in the Supercomputing Center of Wuhan University.

\appendix
{\centering\section*{Appendix}}

\section{More Experimental Results}
\subsection{Effectiveness of EPQM on Bezier Variant Detector}
\begin{figure}[h]
    \centering
    \includegraphics[width=0.6\linewidth]{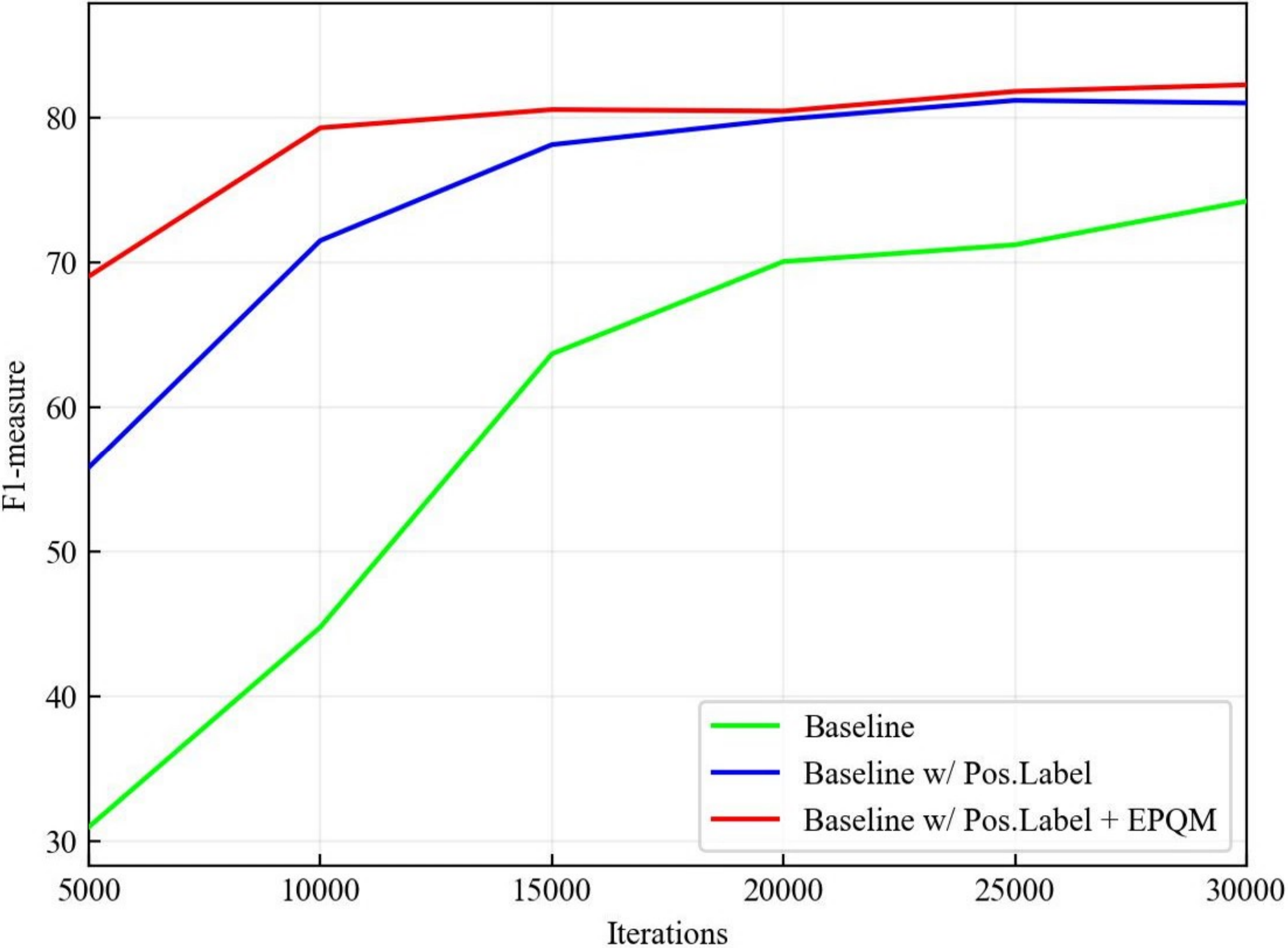}
    \caption{Convergence curves on Rot.Total-Text.}
    \label{fig:bezier_curve}
    \vspace{-5mm}
\end{figure}

In Table 3 of the main paper, we provide the quantitative results which prove the effectiveness of the proposed EPQM on the Bezier variant detector. Here, we additionally show the effectiveness of EPQM on the training convergence in Figure \ref{fig:bezier_curve}. We can find that with EPQM 10\% gains on F1-measure are yielded.

\subsection{Ablations on Circular Convolution}
Here, we conduct ablations about the circular convolution used in our EFSA module. As shown in Table 1, using one layer and 4-neighbourhood, our method achieves the best trade-off between the performance and inference speed. We adopt this setting for all polygon control point version detector in all experiments.

\begin{table}[h]
    \centering
    \resizebox{0.5\hsize}{!}{
    \begin{tabular}{cccc}
         \Xhline{1.2pt}
         Layer &Adj. &F-measure &FPS \\
         \midrule
         1 &2 &85.33 &17.3 \\
         1 &4 &\textbf{86.17} &\textbf{17.3} \\
         1 &6 &85.39 &17.3 \\
         \midrule
         2 &2 &84.57 &16.4 \\
         2 &4 &86.14 &16.4 \\
         2 &6 &85.12 &16.4 \\
         \Xhline{1.2pt}
    \end{tabular}}
    \caption{Ablations on the circular convolution used in the EFSA module. ``Adj.'' represents the number of adjacency elements.}
    \label{tab:ap_circonv}
    \vspace{-5mm}
\end{table}

\begin{table}[h]
    \centering
    \resizebox{\hsize}{!}{
    \begin{tabular}{lcccccccccc}
         \Xhline{1.2pt}
         \multirow{2}{*}{Method} &
         \multirow{2}{*}{Pos.Label} &
         \multirow{2}{*}{EPQM} &
         \multirow{2}{*}{EFSA} &
         \multirow{2}{*}{Rotation} &
         \multicolumn{2}{c}{\textbf{Total-Text}} &
         \multicolumn{2}{c}{\textbf{Rot.Total-Text}} &
         \multicolumn{2}{c}{\textbf{Inverse-Text}} \\
         \cmidrule(lr){6-7} \cmidrule(lr){8-9} \cmidrule(lr){10-11}
         &&&& &F & FPS & F & FPS & F & FPS \\
         \midrule
         $Baseline+$ &\checkmark &\checkmark & & &\textbf{85.80} &\textbf{17.0} &\textbf{73.82} &\textbf{19.0} &\textbf{80.86} &\textbf{18.9} \\
         $Baseline+$ &\checkmark &\checkmark &\checkmark & &85.21 &16.4 &72.74 &18.4 &79.41 &18.2\\
         \midrule
         $Baseline+$ &\checkmark &\checkmark & &\checkmark &\textbf{86.64} &\textbf{17.0} &85.03 &\textbf{19.0} &\textbf{86.20} &\textbf{18.9} \\
         $Baseline+$ &\checkmark &\checkmark &\checkmark &\checkmark &86.26 &16.4 &\textbf{85.21} &18.4 &85.46 & 18.2 \\
         \Xhline{1.2pt}
    \end{tabular}}
    \caption{Ablations on the EFSA used in Bezier control point version detector.}
    \label{tab:appendix_ablation_bezier}
    \vspace{-5mm}
\end{table}

\begin{figure}[!t]
    \centering
    \subcaptionbox{Cared instances counts distribution}{\includegraphics[width=0.9\linewidth]{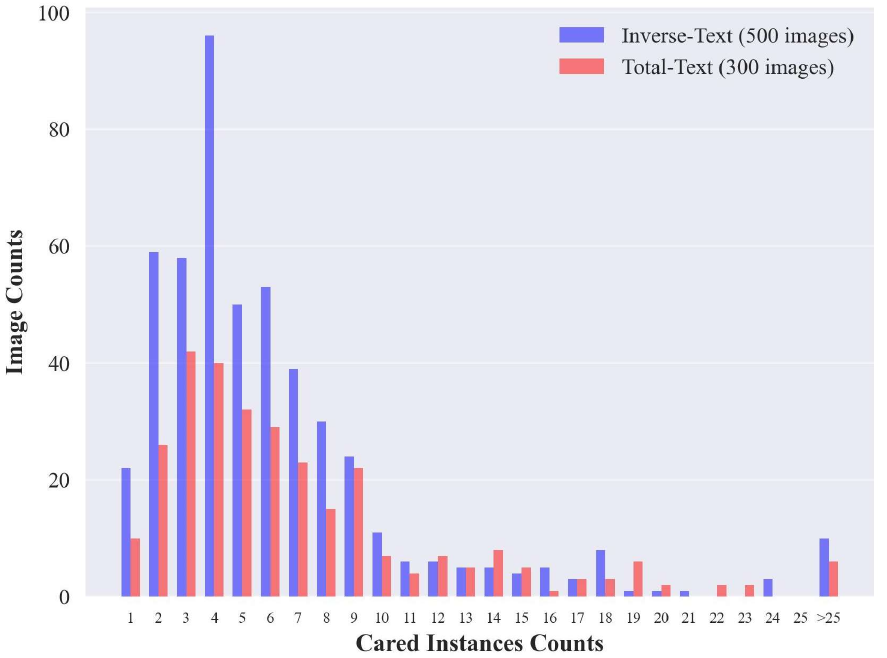}}
    \subcaptionbox{Text length distribution}{\includegraphics[width=0.9\linewidth]{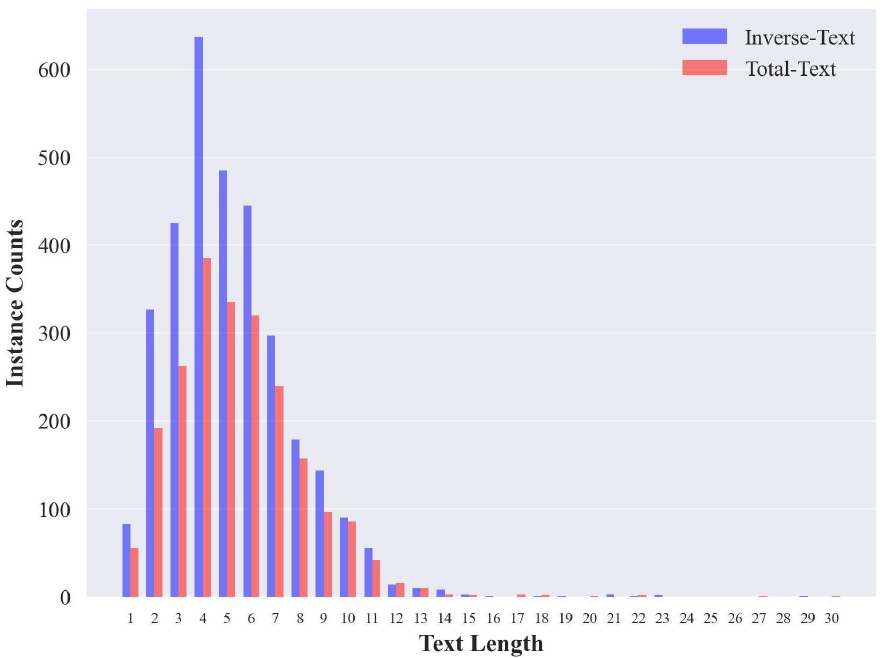}}
    \subcaptionbox{Character frequency.}{\includegraphics[width=0.9\linewidth]{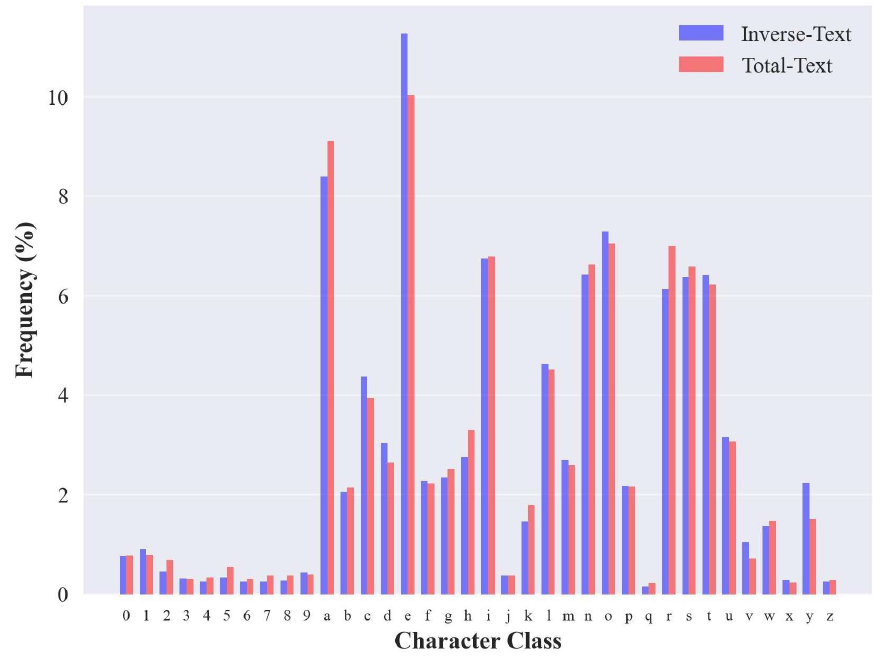}}
    \caption{Statistical analysis on Inverse-Text compared with Total-Text. (a) The number of images with different cared text instances counts. (b) The number of cared instances with different text length. (c) Frequency distribution of different characters.}
    \label{fig:appendix_statistic}
\end{figure}

\subsection{Ablations on EFSA}
For Bezier variant detector, in addition to the Table 3 of the main paper, we present the ablations on EFSA in this subsection. Quantitative results are reported in Table \ref{tab:appendix_ablation_bezier}. We find EFSA impedes the overall performance of Bezier control point version detector. Bezier curve control points do not always form in circular shape and sometimes they are far apart. We conjecture that the circular convolution in EFSA introduces noise to the prediction of Bezier curve control points.

\section{More Details about Inverse-Text}
In this section, we provide more information about Inverse-Text. Specifically, more statistical details are reported in Sec.~\ref{subsec:more_details} and more sample images are displayed in Sec.~\ref{subsec:more_samples}.

\subsection{Statistical Details}
\label{subsec:more_details}
Inverse-Text consists of 500 testing images with word-level polygon annotations. In existing available arbitrary-shape scene text test sets, the same word-level polygon annotations are provided in Total-Text \cite{ch2020total}. Different text-line level polygon annotations are provided in CTW1500 \cite{liu2019curved}. Hence, we only compared the statistical results of Inverse-Text with Total-Text, as shown in Figure \ref{fig:appendix_statistic}. Only Latin characters and digits are considered for recognition. As can be seen, the distribution of the cared instances counts, text length, and character classes are similar between Inverse-Text and Total-Text. Most testing images contain no more than 15 cared text instances which are involved in evaluation. Moreover, the distribution of text length and character frequency are roughly in line with the English corpus.

\subsection{More Samples}
\label{subsec:more_samples}
\begin{figure}[!t]
    \centering
    \includegraphics[width=\linewidth]{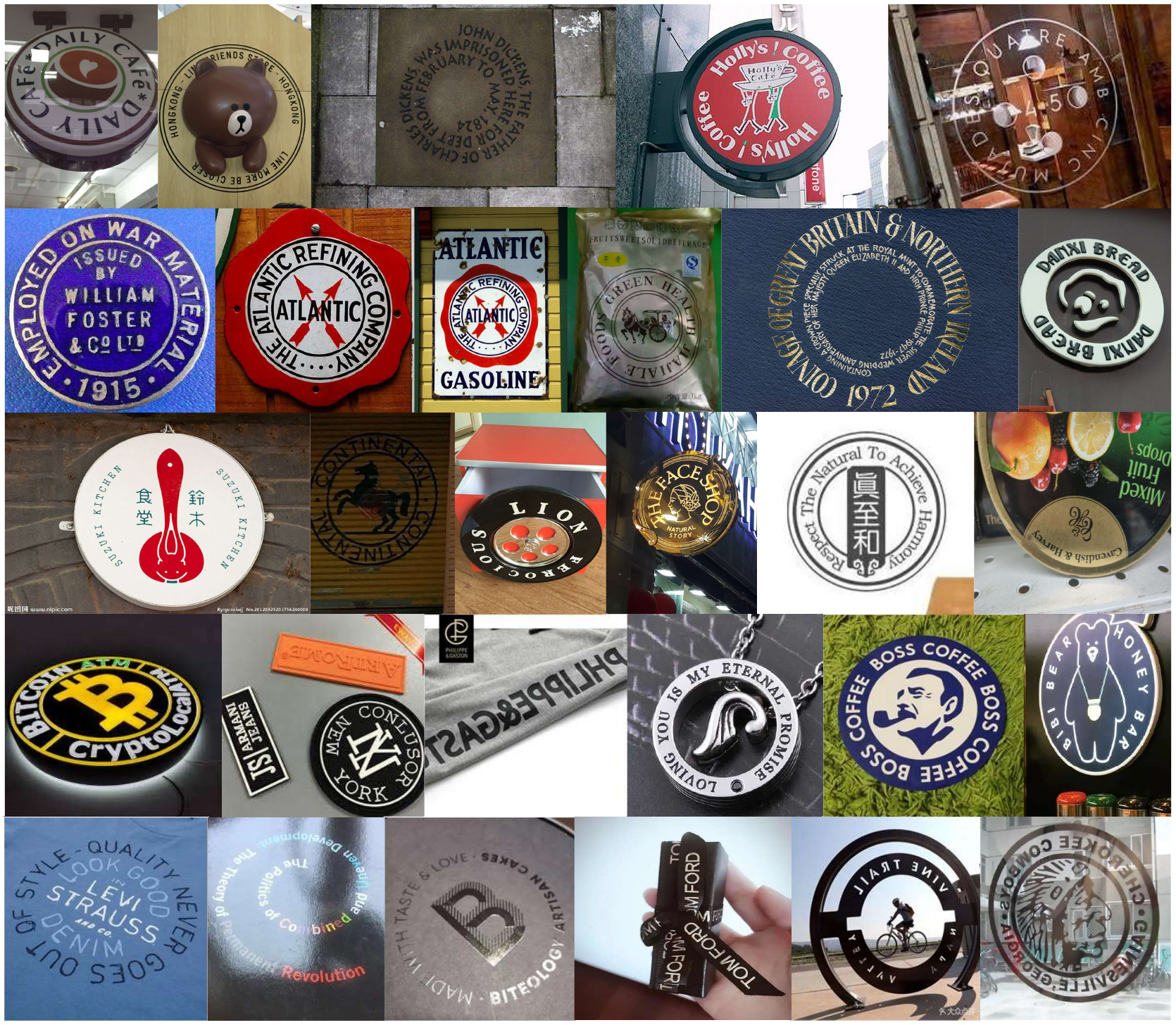}
    \caption{More samples of Inverse-Text.}
    \label{fig:appendix_inversetext}
\end{figure}

More samples are shown in Figure \ref{fig:appendix_inversetext}.

\subsection{Results of Some Spotters}
\begin{figure}[!t]
    \centering
    \subcaptionbox{ABCNet-v2}{\includegraphics[width=\linewidth]{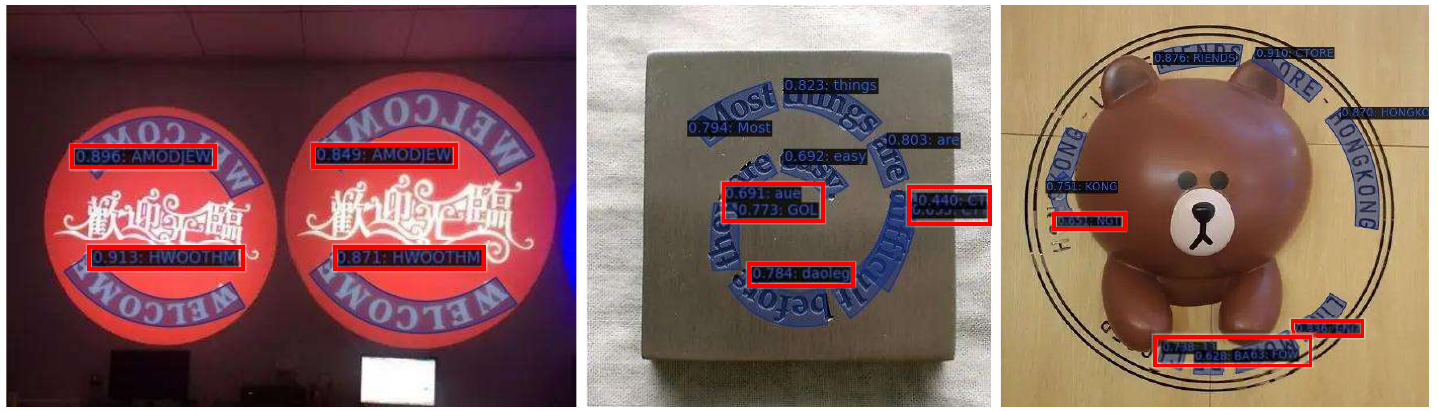}}
    \subcaptionbox{TESTR}{\includegraphics[width=\linewidth]{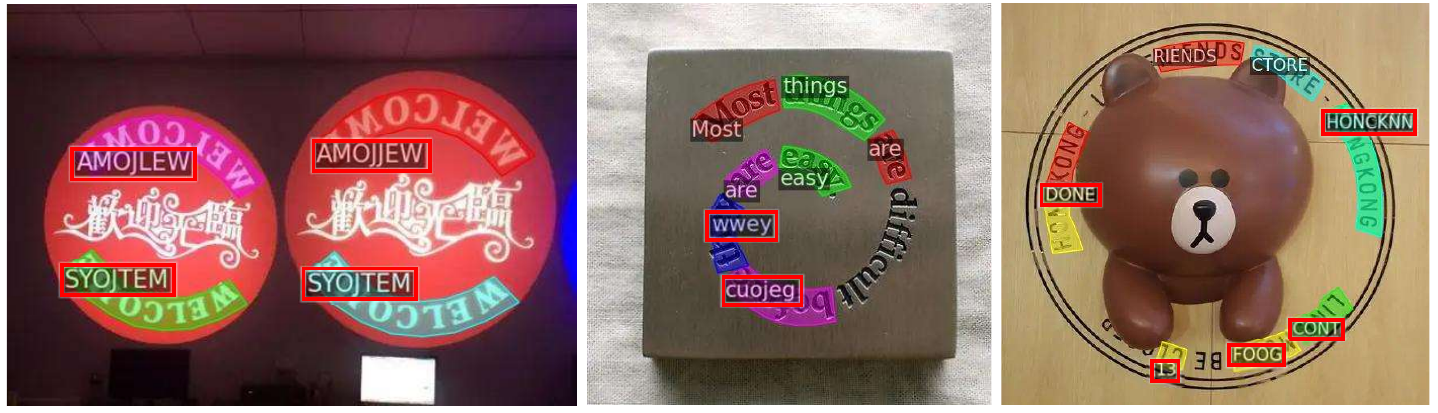}}
    \subcaptionbox{SwinTextSpotter}{\includegraphics[width=\linewidth]{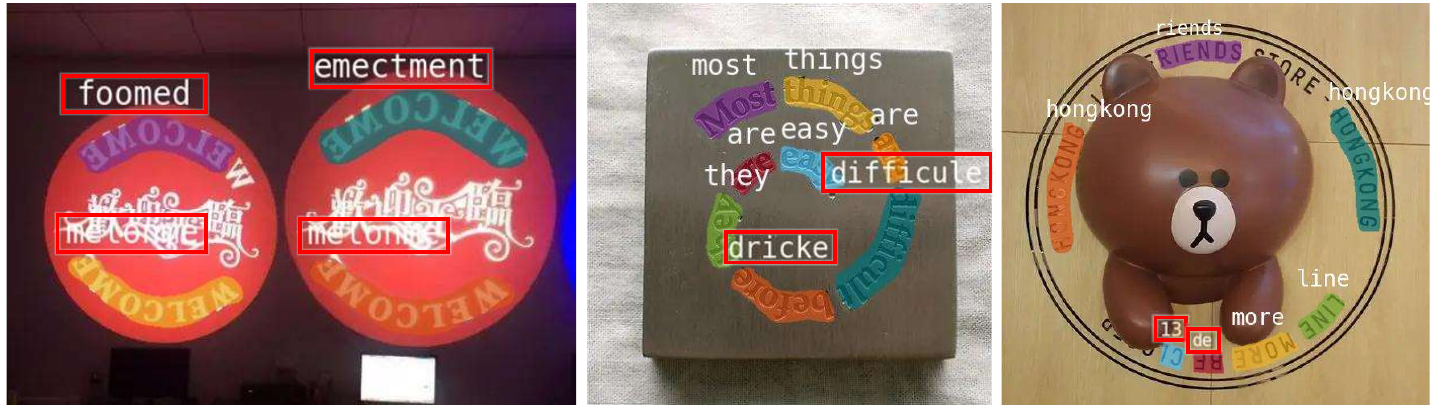}}
    \caption{Some qualitative results on Inverse-Text.}
    \label{fig:appendix_spotters_vis}
\end{figure}

We utilize some previous state-of-the-art spotters to get quantitative and qualitative results on Inverse-text, as shown in Table \ref{tab:appendix_e2e_inversetext} and Figure \ref{fig:appendix_spotters_vis}. We select ABCNet-v2 \cite{9525302} and TESTR \cite{zhang2022text} representing the methods which predict control points. We select SwinTextSpotter \cite{huang2022swintextspotter} representing the segmentation-based spotting methods. Officially released model weights trained on Total-Text \cite{ch2020total} are adopted for direct evaluation. We can see that the spotters struggle to correctly recognize inverse-like texts because of the unconventional reading order. It indicates that there is still a lot of room for improvement towards the more advanced and more robust end-to-end spotter. We hope that our preliminary attempt in data, \textit{i.e.,} filling the gap of lacking inverse-like texts in existing benchmarks, can also inspire and facilitate future researches on text spotting.

\begin{table}[t!]
    \centering
    \resizebox{\hsize}{!}{
    \begin{tabular}{lccc cc}
         \Xhline{1.2pt}
         \multirow{2}{*}{Method} &
         \multirow{2}{*}{P} &
         \multirow{2}{*}{R} &
         \multirow{2}{*}{F} &
         \multicolumn{2}{c}{End-to-End} \\
         \cmidrule(lr){5-6}
         & & & &None &Full \\
         \midrule
         ABCNet-v2 \cite{9525302} &82.0 &70.2 &75.6 &34.5 &47.4 \\
         TESTR \cite{zhang2022text} &83.1 &67.4 &74.4 &34.2 &41.6\\
         SwinTextSpotter \cite{huang2022swintextspotter} &94.5 &85.8 &89.9 &55.4 &67.9 \\
         \Xhline{1.2pt}
    \end{tabular}}
    \caption{Results of spotters on Inverse-Text. ``None'' and ``Full'' indicate end-to-end results without using lexicon and with lexicon, respectively.}
    \label{tab:appendix_e2e_inversetext}
\end{table}

\bibliography{aaai23}

\end{document}